\documentclass[conference]{IEEEtran}
\IEEEoverridecommandlockouts
\usepackage{cite}
\usepackage{amsmath,amssymb,amsfonts}
\usepackage{algorithmic}
\usepackage{graphicx}
\usepackage{textcomp}
\usepackage{xcolor}
\usepackage{makecell}
\usepackage{float,lscape}
\usepackage{tabularx}
\def\BibTeX{{\rm B\kern-.05em{\sc i\kern-.025em b}\kern-.08em
    T\kern-.1667em\lower.7ex\hbox{E}\kern-.125emX}}
    
\begin{document}
\title{Transforming ECG Diagnosis: An In-depth Review of Transformer-based Deep Learning Models in Cardiovascular Disease Detection
}

\author{\IEEEauthorblockN{Zibin Zhao}
\IEEEauthorblockA{\textit{Department of Chemical and Biological Engineering} \\
\textit{Hong Kong University of Science and Technology}\\
Hong Kong, P. R. China \\
E-mail: zibin.zhao@connect.ust.hk}

}

\maketitle

\begin{abstract}
The emergence of deep learning has significantly enhanced the analysis of electrocardiograms (ECGs), a non-invasive method that is essential for assessing heart health. Despite the complexity of ECG interpretation, advanced deep learning models outperform traditional methods. However, the increasing complexity of ECG data and the need for real-time and accurate diagnosis necessitate exploring more robust architectures, such as transformers. Here, we present an in-depth review of transformer architectures that are applied to ECG classification. Originally developed for natural language processing, these models capture complex temporal relationships in ECG signals that other models might overlook. We conducted an extensive search of the latest transformer-based models and summarize them to discuss the advances and challenges in their application and suggest potential future improvements. This review serves as a valuable resource for researchers and practitioners and aims to shed light on this innovative application in ECG interpretation. \\
\end{abstract}


\section{Introduction}
The development of deep learning has led to significant breakthroughs in various fields, including healthcare. One area where it has made a particularly profound impact is in the analysis of electrocardiograms (ECGs)\cite{c1, c2}. ECGs are noninvasive tests that measure the electrical activity of the heart and play a critical role in assessing heart health. However, interpreting ECGs requires extensive education and training \cite{c3, c4}. The integration of deep learning into ECG analysis has ushered in a new era of improved accuracy.

In recent years, there has been a surge of research exploring deep learning's potential in ECG diagnosis \cite{c1, c5}. Various architectures, such as Stacked Auto-Encoders (SAE)\cite{c6}, Deep Belief Networks (DBN)\cite{c7}, Convolutional Neural Networks (CNN)\cite{c8}, and Recurrent Neural Networks (RNN)\cite{c9}, have been developed and have shown comparably better performance than manual classifications by experts. However, due to the increasing complexity of ECG data and the need for more accurate and real-time diagnosis, more robust and efficient deep learning architectures are needed.

Transformers, originally designed for natural language processing tasks, have been introduced to ECG classification. Transformers' self-attention mechanism \cite{c10} allows for the consideration of the entire sequence of an ECG signal, potentially capturing complex temporal relationships that other architectures might miss. However, there are few comprehensive reviews on the application of transformer architectures to ECG classification.

This paper aims to provide a detailed overview of the advances and challenges in applying transformer architectures to ECG classification. We will analyze and summarize the technical underpinnings of transformer models, and their application to ECG data in terms of accuracy, efficiency, significance, and potential challenges. Additionally, we will discuss the limitations of the current approaches and the potential improvements to be made on a broader scale for the ECG community in the future. We believe this review will be a valuable resource for researchers and practitioners in the field, shedding light on the novel use of transformer architectures in ECG classification and paving the way for future innovations.

This literature review focuses specifically on transformer-based models in the context of electrocardiogram (ECG) interpretation. While conventional machine learning and other deep learning technologies also play important roles in this field, we will briefly introduce the current advancements but will not be discussing them extensively in this review. This is because there are already many excellent reviews that comprehensively cover these methodologies in the context of ECG analysis\cite{c1,c5,c11,c12}. Our primary discussion and comparative analysis will be reserved for the innovative use of transformer-based models in ECG interpretation. This paper is organized as follows: The current state-of-the-art ECG deep learning models will be summarized in Section 2. Section 3 discusses some novel deployments of transformers in ECG. In Section 4, we present both challenges and opportunities for deep learning in ECG society. Finally, a brief conclusion is drawn in Section 5. 

\section{Current development on ECG with deep learning}

Deep learning has been developed and refined for ECG diagnosis over many years. However, different models use various types of ECG signals in different scenarios. In conventional ECG monitoring, 12-lead ECGs are the gold standard in hospitals for comprehensive diagnosis of various heart diseases\cite{c13}. Single-lead and other types of ECG signals are also important and employed by researchers in different circumstances. It is important to distinguish between them before comparing them all at once.

\subsection{Single-lead vs. multi-lead}

A single-lead ECG records the heart's electrical activity from a single lead or viewpoint. It is less comprehensive than a multi-lead ECG, but simpler, more convenient, and less invasive. It is often used in wearable devices, home monitoring systems. In contrast, a multi-lead ECG records the heart's electrical activity from multiple dimensions, usually 12, but sometimes more. Each lead provides a different perspective on the heart's electrical activity, offering a more comprehensive picture of heart function. Multi-lead ECGs are typically used in clinical settings and are invaluable for diagnosing a wide range of heart conditions\cite{c13}. However, more and more 12-lead ECGs, as a more accurate and reliable indicator of heart health, are being developed and deployed into wearable devices for home use\cite{c14}. Such devices are particularly useful for diagnosing heart diseases like ischemia or left/right bundle branch block, which can only be spotted from 12-lead ECG signals\cite{c15}.  It is obvious that deep learning models that are trained on single-lead ECG data may not perform as well as those trained on multi-lead data due to the reduced complexity and dimensionality of the single-lead data. Although single-lead ECGs are generally easier to train and require fewer computational resources.

\subsection{Static vs. dynamic ECG signal}
ECG signals can be categorized as static or motion ECGs. Static ECGs are collected at rest and are commonly used in clinical settings for routine check-ups or to diagnose heart conditions. They tend to be cleaner and have fewer artifacts because there is less interference from physical movement. Motion, dynamic, or stress ECGs are collected while the individual is moving, such as during exercise or regular daily activities. These ECGs are useful for capturing heart activity under stress or for monitoring heart health during daily life. However, they can be more challenging to interpret due to the potential for noise and artifacts introduced by physical movement\cite{c16}. DL models can be trained on both static and motion ECG data to monitor heart health. However, challenges may arise in motion ECG due to the potential for noise and artifacts in the data. Therefore, techniques such as noise filtering, data augmentation, and robust training methods have been implemented to improve performance\cite{c17}.

\subsection{Current state-of-the-art ECG deep learning models}
In this session, two static and two dynamic ECG DL models that were selected to represent the most recent developments in this area are discussed in Table 1.

CNNs are excellent at capturing local and spatial patterns\cite{c8}, while BiLSTMs capture temporal dependencies in the ECG signal\cite{c18}. Additionally, CNNs can filter out noise and artifacts in ECG signals, enabling robust diagnosis even in the presence of interference. BiLSTM models process ECG signals in both forward and backward directions, providing richer context and improving the model's ability to recognize complex arrhythmias.

When training on static ECGs, variations on CNN-based models are frequently used by researchers, as they are adept at extracting spatial information from ECG signals or images. Furthermore, 12-lead ECG signals are increasingly common in static ECG DL model training \cite{c19} because they provide the model with more vector information to learn from in a more comprehensive manner, leading to more accurate and reliable results.

Dynamic ECG models, however, generally employ single-lead ECG due to device limitations in acquiring ECG signals from an exercising object. Only stress ECG tests at hospitals are available for exercise 12-lead ECG. \cite{c20} Additionally, signals are susceptible to noise due to human movement, which can cause baseline drift or physical leads to hang over and move, resulting in significant noise artifacts in the signals and thus affecting the model's performance.

\renewcommand{\arraystretch}{1.8}
\begin{table*}[h]
  \centering
  \caption{Advanced ECG Diagnostic with Deep Learning Summary}
  \begin{tabular}{lllllll}
    \hline 
    \textbf{Authors} & \textbf{Application}  & \textbf{Type} & \textbf{Lead(s)} & \textbf{DL algorithm} & \textbf{Dataset} & \textbf{Remarks}  \\ 
    \hline 
    Ribeiro et al.\cite{c21}  & 
    \makecell[l]{Six types of abnormalities \\ classification} & Static & 12 & Residual CNN & TNMG Brazil & \makecell[l]{Cardiologists validation \\ with high specificity 99\% } \\
    \hline
    Sangha et al.\cite{c22}  & \makecell[l]{Multilabel abnormalities \\ diagnosis on ECG images} & Static & 12 & EfficientNet CNN & TNMG Brazil & \makecell[l]{Grad-CAM improves \\ Interpretability; \\ One hidden lablel for gender} \\
    \hline
    Guan et al.\cite{c23}  & HRV detection & Dynamic & 1 & \makecell[l]{SMOTE \\ BiLSTM + Attention} & \makecell[l]{Self-collected \\database} & \makecell[l]{Combine features of \\acceleration, ECG and \\angular velocity; Multi-sensor \\feature-level fusion}  \\
    \hline
    Lee et al.\cite{c24}  & \makecell[l]{Atrial fibrillation detection\\ with P-waves location} & Dynamic & 1 & \makecell[l]{CNN \\ BiLSTM + Attention} & \makecell[l]{Kaohsiung Medical \\ University Hospital} & \makecell[l]{CNN for P-waves location; \\ BiLSTM for AF prediction} \\
    \hline
  \end{tabular}
  \end{table*}
\hfill

\section{ECG with Transformer}

\subsection{Transformer}
The Transformer is a deep learning model designed for sequence-to-sequence tasks in natural language processing (NLP), such as translation and text summarization. It was introduced in the 2017 paper "Attention is All You Need" by Vaswani et al.\cite{c10} The Transformer's architecture (see Fig. 1) differs from that of other sequence processing models, such as Recurrent Neural Networks (RNNs) or Long Short-Term Memory Networks (LSTMs). Transformers allow for more efficient parallelization during training, leading to faster and more scalable models by doing away with the sequential computation inherent to RNNs and LSTMs. This makes Transformers a promising model in many areas\cite{c25, c26}.

\subsection{Transformer with ECG}
Electrocardiogram (ECG) signals and images often contain significant spatio-temporal data in their raw form or transformed spectrographic representation. Researchers have extensively used machine learning and deep learning architectures to extract and understand latent features directly from ECG signals, leading to the development of highly effective models for ECG classification tasks\cite{c27,c28}. However, cardiac conditions often manifest episodically\cite{c29}, resulting in disease morphologies that may not frequently appear in ECG signals. Therefore, conventional model architecture, even with comprehensive feature exploration, may overlook highly specific disease-related features within the total signal length. Additionally, detailed feature extraction at every single time step could potentially neglect broader morphological patterns, leading to time-consuming procedures and increased computational demand.

The self-attention mechanism, also known as Transformer attention or scaled dot-product attention, is a key innovation of the Transformer model. This mechanism helps the model assess and prioritize different elements within the input sequence, capturing relationships between these elements independent of their sequential positions\cite{c30}. This attention mechanism has proven successful across various domains, including computer vision and large language models, by enabling the model to highlight important features. Increasingly, researchers have utilized and adapted the self-attention transformer for ECG applications, achieving promising outcomes.

Regarding the use of transformers on ECG signals, the following advantages and disadvantages may be observed:
\textbf{PROS:}
\begin{itemize}
    \item Transformer is capable of processing various arbitrary-length single or 12-lead ECG signals with different input lengths, making them flexible and adaptable for a wide range of ECG data\cite{c31}.
    \item Transformer when combined with Convolutional Neural Networks (CNNs), can handle both spatial and temporal information in ECG signals, enhancing the model's ability to detect abnormalities and arrhythmias \cite{c1, c32}.
    \item The sequence-to-sequence structure enables the model to output the positions of all heartbeats within the input ECG segments while also predicting their classes at the same time.
\end{itemize}

\noindent
\textbf{CONS:}
\begin{itemize}
    \item Transformer models are complex and computationally intensive, which can be a disadvantage for real-time or on-device applications (wearable) where computational resources are limited.
    \item Transformers often require a large amount of labeled data for training. The collection and annotation of such data can be time-consuming and expensive.
    \item Risk of overfitting with transformers, especially when dealing with small or imbalanced datasets.
    \item Interpretability can be a challenge with transformer models. While they can make accurate predictions, understanding why they made a certain prediction is not straightforward, which could be problematic in a healthcare setting where interpretability is often necessary for clinicians to trust and act on model predictions \cite{c33}. 
\end{itemize}

\begin{figure}[h!]
\centerline{\includegraphics[width=50mm]{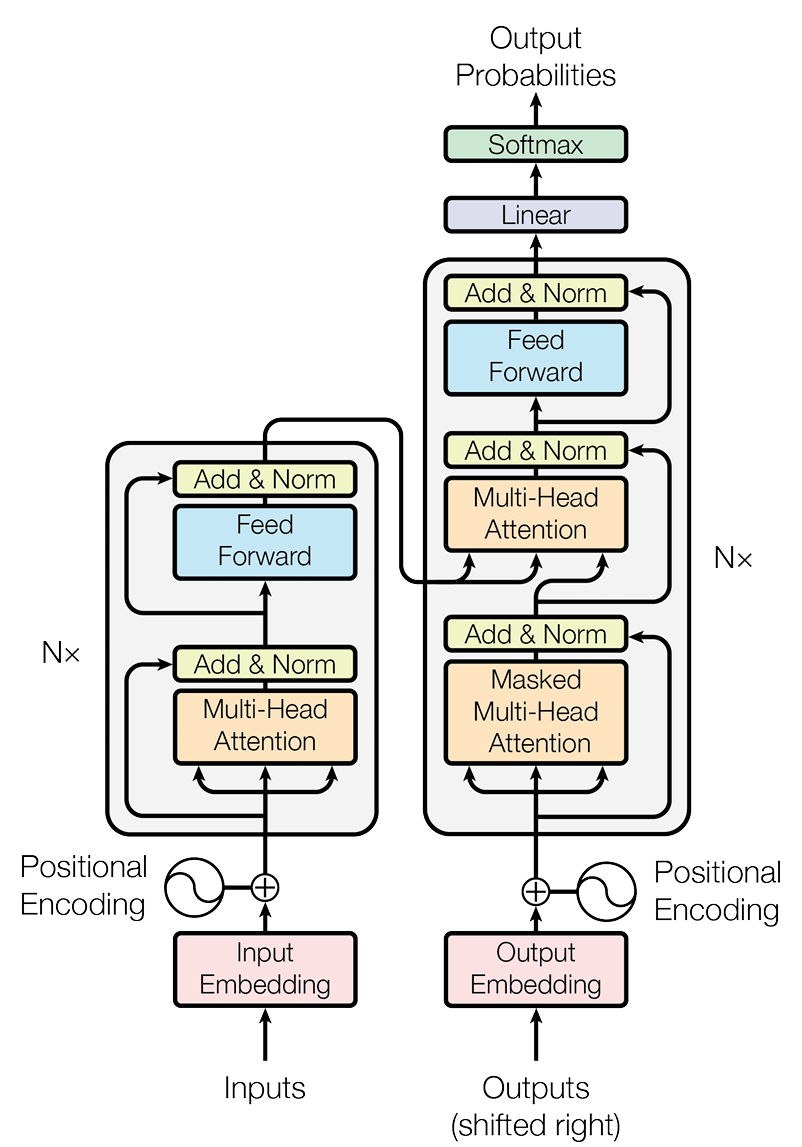}}
\caption{The transformer model architecture from Vaswani et al. \cite{c10}}
\label{fig}
\end{figure}

In Table 2, we summarize the latest research on deploying transformer-based ECG diagnosis models. \\

Many studies utilize the MIT-BIH Database for their research, demonstrating its widespread acceptance and robustness for arrhythmia-related studies, as well as for other types of cardiac diseases\cite{c34}. A variety of architectures merge Convolutional Neural Networks (CNNs) with transformer models, demonstrating the efficacy of combining these two powerful deep learning techniques to capture both local features and long-range dependencies in ECG signals. This combination can enhance the performance of arrhythmia detection and classification tasks.

\newpage
\begin{landscape}
\newcolumntype{l}{>{\raggedright\arraybackslash}X} 
\begin{table}[ht!]
\begin{tabularx}{\linewidth}{X*{8}{l}}
\hline
\textbf{Author} & \textbf{Title} & \textbf{Classification Types} & \textbf{Lead(s)} & \textbf{Database} & \textbf{Architecture} & \textbf{Significance} & \textbf{Accuracy (average)}\\
\hline
Liu et al. \cite{c35}  & CRT-Net: A generalized and scalable framework for the computer-aided diagnosis of Electrocardiogram signals & AAMI Standard & 12 & \parbox[t]{2cm}{MIT-BIH \\ CPSC 2018} & Conv, RNN, Transformer CRT-Net & Bi-connectivity extracts ECG signals from  2D images; cloud-based system; multiple databases validation & \parbox[t]{2cm}{Accuracy: 99.6\% \\F1 score: 99.6\%} \\
Yang et al. \cite{c36} & Automated diagnosis of atrial fibrillation using ECG component-aware transformer & Binary classification of Atrial fibrillation & 12 & ShaoXing database & MLP + Component-aware Transformer (CAT) & Segment and tokenize the ECG features using 1D U-net as input & \parbox[t]{3cm}{Accuracy: 98.23\% \\F1 score: 85.13\%} \\
Chen at al. \cite{c37} & Detection and Classification of Cardiac Arrhythmia by a Challenge-Best Deep Learning Neural Network Model & 8 classes Cardiac arrhythmia & 12 & CPSC 2018 & CNN + BiGRU + Attention & Large ECG dataset; ensemble model for both 1-/12-leads ECG & \parbox[t]{2cm}{Accuracy: 99.5\% \\F1 score: 84.0\%} \\
Meng et al. \cite{c38} & Enhancing dynamic ECG heartbeat classification with lightweight transformer model & PVCs \& SPBs & 1 & \parbox[t]{2cm}{Self-collected \\dataset} & CNN embedding + LightConv attention & \parbox[t]{3cm}{Lightweight fussing \\transformer with \\ ablation study; \\ Noise removal} & \parbox[t]{3cm}{Accuracy: 99.32\% \\F1 score: 93.63\%} \\
Yan et al. \cite{c39} & Fusing Transformer Model with Temporal Features for ECG Heartbeat Classification & AAMI (exclude class Q) & 1 & MIT-BIH & Fussing Transformer + RR interval features embedding & Encoder only; simple and paralleled structure; fusing handcrafted features with SMOTE; extensive validation & \parbox[t]{3cm}{Accuracy: 99.62\% \\F1 score: 94.56\%} \\
Hu et al. \cite{c40} & A transformer-based deep neural network for arrhythmia detection using continuous ECG signals & AAMI; 8 classes Arrhythmia & 1 & MIT-BIH arrhythmia; MIT-BIH Atrial fibrillation & CNN + Transformer & Two models on two databases; convert ECG into object detection task & \parbox[t]{3cm}{Accuracy: 99.23\% \\F1 score: 99.23\% \\ (highest class)} \\
Bing et al. \cite{c41} & Electrocardiogram classification using TSST-based spectrogram and ConViT & AAMI & 1 & MIT-BIH & TSST + ConViT & Segmentation and time-reassigned synchrosqueezing transform ECG as model input & \parbox[t]{4cm}{Accuracy: 99.5\% \\F1 score: 94.0\%} \\
Le et al. \cite{c42} & Multi-module Recurrent Convolutional Neural Network with Transformer Encoder for ECG Arrhythmia Classification & AAMI & 1 & MIT-BIH Arrythmia & LSTM-Transformer & Multi-module fusion to enhance temporal dependencies between 1D segments and 2D spectrograms & \parbox[t]{3cm}{Accuracy: 98.29\% \\F1 score: 99.14\% \\ (highest class)} \\
\hline
\end{tabularx}
\end{table}
\end{landscape}

Despite these shared aspects, differences in study designs and methodologies underscore the diverse strategies applied in this field. For instance, the classification types varied across the studies, with some focused on specific arrhythmias like PVCs and SPBs or Atrial fibrillation, while others adopted a broader scope by considering multiple classes defined by the AAMI standard\cite{c43}. The transformer model's design and application also varied significantly. Some studies utilized a more lightweight or simplified version of the transformer model, while others incorporated more complex features, such as attention mechanisms or custom embeddings. For example, Meng et al. \cite{c38} employed a lightweight transformer model with a noise removal mechanism, while Yan and their team \cite{c39} utilized the encoder part of the transformer only and fused it with handcrafted features. Additionally, the integration of other deep learning models or techniques with the transformer model was observed. For instance, Chen et al. \cite{c37} incorporated a BiGRU with attention, and Hao and Nugroho \cite{c44} proposed a spatio-temporal approach using a Conv-Transformer. 

As mentioned earlier, different types of ECG and the number of leads can play various roles in different scenarios. To the best of our knowledge, transformer-based models are only employed with static ECG data, but researchers focus differently on the level and type of ECG data complexity, with some focusing on single-lead ECG data and others utilizing 12-lead ECG data. Additionally, the performance of the models, as indicated by their F1 scores and accuracy rates, is provided for comparison.

\section{Discussion}
After conducting research on conventional and transformer-based deep learning models, it is clear that deep learning has been a well-established area of study in ECG diagnosis that has continued to grow over the years. The disparities in Table 2 highlight the need for further investigation and comparison of different transformer-based models and techniques for ECG analysis, taking into account the nature of the ECG data and the specific classification tasks. Despite its success and maturity, there are still potential challenges and opportunities that will be discussed below.

\subsection{Dataset and the imbalance challenge}
Opting for a small or older database might seem practical or convenient, but it may not provide a broad enough perspective, particularly when examining the relationships among various leads. Moreover, an approach that only considers a single lead tends to overlook the interconnectedness between leads, potentially missing out on important diagnostic information.

Additionally, there are certain challenges associated with handling ECG data, particularly the imbalance problem. Some researchers attempt to tackle this using the Synthetic Minority Over-sampling Technique (SMOTE), but it often falls short of producing desired outcomes. In fact, inserting null values as a balancing technique may lead to increased noise, thereby affecting the overall data quality.

\subsection{Model architecture, transparency, and interpretability}
One common point regarding model architectures is the fusion of Convolutional Neural Networks (CNNs) with transformer models. This combination leverages the strength of CNNs in extracting local features from ECG data and transformers in capturing long-range dependencies. Looking ahead, there are several ways to improve the current models. One approach involves developing lighter model architectures that can be implemented on wearable ECG devices using pre-trained weights. Another approach involves enhancing the processing of dynamic ECG signals and improving disease classification, whether for single-lead or 12-lead ECGs.

A reasonable ablation study is necessary in the current era of modified models. Such studies can help understand the contribution of each component of the model. Additionally, the validation of these models should not only focus on their performance but also their interpretability. More emphasis should be placed on exploring the interrelationships between each lead during the embedding process to enhance the model's diagnostic capability.

A key aspect of these advancements should be increasing the model's transparency during training. This can be achieved by integrating more visual-enabled technologies, such as Grad-CAM, to enhance model interpretability. This is particularly beneficial in clinical settings where physicians can leverage this transparency for better decision-making. Indeed, the ongoing trend in ECG diagnosis using deep learning emphasizes the importance of model interpretability. The ideal scenario involves validation by cardiologists to ensure the development of a robust and convincing diagnostic system.

\subsection{Benchmarking and evaluation metric}
As mentioned earlier, a good database may not only help in training but also as a good reference in benchmarking. Developing more standardized benchmarks and evaluation metrics would greatly facilitate the comparison and validation of these models in the future. A key benchmark could be the adoption of uniform datasets for model training and testing. For instance, many researchers have adopted the MIT-BIH Database as a standard benchmark. However, such databases have been utilized in the past 20 years and may not provide refreshed and useful inference in today's clinical settings. Utilizing a consistent database across various studies would ensure that the models are tested on the same ground, enabling a more reliable comparison of their performance.

As for evaluation metrics, accuracy, and F1 score are commonly used. Accuracy measures the proportion of correct predictions out of all predictions, while the F1 score is the harmonic mean of precision and recall, providing a balanced measure of a model's performance, especially on imbalanced datasets. These metrics should be consistently reported for all classes in multiclass classification tasks, not just the average or the highest-performing class. This would provide a comprehensive view of a model's performance across all classes. Other metrics such as sensitivity, specificity, and Area Under the Receiver Operating Characteristic Curve (AUC-ROC) could also be considered for a more nuanced evaluation.

It should be noted that while standardization can enhance comparability, innovation should be encouraged. Therefore, while these benchmarks and metrics could serve as a baseline, studies should not be confined to them and should continue to explore novel architectures and evaluation methods.

\section{Conclusion}
In this paper, the application of transformer-based deep learning architectures for electrocardiogram (ECG) classification is studied and summarized. Given the increasing complexity of ECG data and the need for precise, real-time diagnosis, the exploration of efficient and robust deep learning architectures like transformers has become essential. We provided a detailed overview of the current advances, while identifying the challenges associated with applying transformer architectures to ECG classification. Furthermore, we discussed potential improvements to further enhance the performance of these models in ECG interpretation. By discussing several commonalities and distinctions in the application of transformer-based models for ECG diagnosis, we hope to offer potential directions for future research in this critical area of healthcare technology.

\bibliographystyle{IEEEtran}
\bibliography{reference}

\begin{thebibliography}{10}
\providecommand{\url}[1]{#1}
\csname url@samestyle\endcsname
\providecommand{\newblock}{\relax}
\providecommand{\bibinfo}[2]{#2}
\providecommand{\BIBentrySTDinterwordspacing}{\spaceskip=0pt\relax}
\providecommand{\BIBentryALTinterwordstretchfactor}{4}
\providecommand{\BIBentryALTinterwordspacing}{\spaceskip=\fontdimen2\font plus
\BIBentryALTinterwordstretchfactor\fontdimen3\font minus
  \fontdimen4\font\relax}
\providecommand{\BIBforeignlanguage}[2]{{%
\expandafter\ifx\csname l@#1\endcsname\relax
\typeout{** WARNING: IEEEtran.bst: No hyphenation pattern has been}%
\typeout{** loaded for the language `#1'. Using the pattern for}%
\typeout{** the default language instead.}%
\else
\language=\csname l@#1\endcsname
\fi
#2}}
\providecommand{\BIBdecl}{\relax}
\BIBdecl

\bibitem{c1}
S.~Somani, A.~J. Russak, F.~Richter, S.~Zhao, A.~Vaid, F.~Chaudhry, J.~K.
  De~Freitas, N.~Naik, R.~Miotto, G.~N. Nadkarni \emph{et~al.}, ``Deep learning
  and the electrocardiogram: review of the current state-of-the-art,'' \emph{EP
  Europace}, vol.~23, no.~8, pp. 1179--1191, 2021.

\bibitem{c2}
N.~Strodthoff, P.~Wagner, T.~Schaeffter, and W.~Samek, ``Deep learning for ecg
  analysis: Benchmarks and insights from ptb-xl,'' \emph{IEEE Journal of
  Biomedical and Health Informatics}, vol.~25, no.~5, pp. 1519--1528, 2021.

\bibitem{c3}
S.-Y. Oh, D.~A. Cook, P.~W. Van~Gerven, J.~Nicholson, H.~Fairbrother, F.~W.
  Smeenk, and M.~V. Pusic, ``Physician training for electrocardiogram
  interpretation: A systematic review and meta-analysis,'' \emph{Academic
  Medicine}, vol.~97, no.~4, pp. 593--602, 2022.

\bibitem{c4}
C.~Breen, G.~Kelly, and W.~Kernohan, ``Ecg interpretation skill acquisition: A
  review of learning, teaching and assessment,'' \emph{Journal of
  electrocardiology}, vol.~73, pp. 125--128, 2022.

\bibitem{c5}
X.~Liu, H.~Wang, Z.~Li, and L.~Qin, ``Deep learning in ecg diagnosis: A
  review,'' \emph{Knowledge-Based Systems}, vol. 227, p. 107187, 2021.

\bibitem{c6}
\BIBentryALTinterwordspacing
S.~Nurmaini, A.~Darmawahyuni, A.~N. Sakti~Mukti, M.~N. Rachmatullah,
  F.~Firdaus, and B.~Tutuko, ``Deep learning-based stacked denoising and
  autoencoder for ecg heartbeat classification,'' \emph{Electronics}, vol.~9,
  no.~1, 2020. [Online]. Available:
  \url{https://www.mdpi.com/2079-9292/9/1/135}
\BIBentrySTDinterwordspacing

\bibitem{c7}
M.~Huanhuan and Z.~Yue, ``Classification of electrocardiogram signals with deep
  belief networks,'' in \emph{2014 IEEE 17th International Conference on
  Computational Science and Engineering}.\hskip 1em plus 0.5em minus
  0.4em\relax IEEE, 2014, pp. 7--12.

\bibitem{c8}
U.~B. Baloglu, M.~Talo, O.~Yildirim, R.~San~Tan, and U.~R. Acharya,
  ``Classification of myocardial infarction with multi-lead ecg signals and
  deep cnn,'' \emph{Pattern recognition letters}, vol. 122, pp. 23--30, 2019.

\bibitem{c9}
S.~Singh, S.~K. Pandey, U.~Pawar, and R.~R. Janghel, ``Classification of ecg
  arrhythmia using recurrent neural networks,'' \emph{Procedia computer
  science}, vol. 132, pp. 1290--1297, 2018.

\bibitem{c10}
A.~Vaswani, N.~Shazeer, N.~Parmar, J.~Uszkoreit, L.~Jones, A.~N. Gomez,
  {\L}.~Kaiser, and I.~Polosukhin, ``Attention is all you need,''
  \emph{Advances in neural information processing systems}, vol.~30, 2017.

\bibitem{c11}
Z.~Ebrahimi, M.~Loni, M.~Daneshtalab, and A.~Gharehbaghi, ``A review on deep
  learning methods for ecg arrhythmia classification,'' \emph{Expert Systems
  with Applications: X}, vol.~7, p. 100033, 2020.

\bibitem{c12}
S.~Hong, Y.~Zhou, J.~Shang, C.~Xiao, and J.~Sun, ``Opportunities and challenges
  of deep learning methods for electrocardiogram data: A systematic review,''
  \emph{Computers in biology and medicine}, vol. 122, p. 103801, 2020.

\bibitem{c13}
K.~E. Sandau, M.~Funk, A.~Auerbach, G.~W. Barsness, K.~Blum, M.~Cvach,
  R.~Lampert, J.~L. May, G.~M. McDaniel, M.~V. Perez \emph{et~al.}, ``Update to
  practice standards for electrocardiographic monitoring in hospital settings:
  a scientific statement from the american heart association,''
  \emph{Circulation}, vol. 136, no.~19, pp. e273--e344, 2017.

\bibitem{c14}
C.-C. Hsu, B.-S. Lin, K.-Y. He, and B.-S. Lin, ``Design of a wearable 12-lead
  noncontact electrocardiogram monitoring system,'' \emph{Sensors}, vol.~19,
  no.~7, p. 1509, 2019.

\bibitem{c15}
M.~J. London, M.~Hollenberg, M.~G. Wong, L.~Levenson, J.~F. Tubau, W.~Browner,
  and D.~T. Mangano, ``Intraoperative myocardial ischemia: localization by
  continuous 12-lead electrocardiography.'' \emph{Anesthesiology}, vol.~69,
  no.~2, pp. 232--241, 1988.

\bibitem{c16}
D.~Tong, K.~Bartels, and K.~Honeyager, ``Adaptive reduction of motion artifact
  in the electrocardiogram,'' in \emph{Proceedings of the Second Joint 24th
  Annual Conference and the Annual Fall Meeting of the Biomedical Engineering
  Society][Engineering in Medicine and Biology}, vol.~2.\hskip 1em plus 0.5em
  minus 0.4em\relax IEEE, 2002, pp. 1403--1404.

\bibitem{c17}
S.~Nayak, M.~Soni, D.~Bansal \emph{et~al.}, ``Filtering techniques for ecg
  signal processing,'' \emph{International Journal of Research in Engineering
  \& Applied Sciences}, vol.~2, no.~2, pp. 671--679, 2012.

\bibitem{c18}
X.~Xu, S.~Jeong, and J.~Li, ``Interpretation of electrocardiogram (ecg) rhythm
  by combined cnn and bilstm,'' \emph{Ieee Access}, vol.~8, pp.
  125\,380--125\,388, 2020.

\bibitem{c19}
A.~Natarajan, Y.~Chang, S.~Mariani, A.~Rahman, G.~Boverman, S.~Vij, and
  J.~Rubin, ``A wide and deep transformer neural network for 12-lead ecg
  classification,'' in \emph{2020 Computing in Cardiology}.\hskip 1em plus
  0.5em minus 0.4em\relax IEEE, 2020, pp. 1--4.

\bibitem{c20}
P.~Gamble, H.~McManus, D.~Jensen, and V.~Froelicher, ``A comparison of the
  standard 12-lead electrocardiogram to exercise electrode placements,''
  \emph{Chest}, vol.~85, no.~5, pp. 616--622, 1984.

\bibitem{c21}
A.~H. Ribeiro, M.~H. Ribeiro, G.~M. Paix{\~a}o, D.~M. Oliveira, P.~R. Gomes,
  J.~A. Canazart, M.~P. Ferreira, C.~R. Andersson, P.~W. Macfarlane,
  W.~Meira~Jr \emph{et~al.}, ``Automatic diagnosis of the 12-lead ecg using a
  deep neural network,'' \emph{Nature communications}, vol.~11, no.~1, p. 1760,
  2020.

\bibitem{c22}
V.~Sangha, B.~J. Mortazavi, A.~D. Haimovich, A.~H. Ribeiro, C.~A. Brandt, D.~L.
  Jacoby, W.~L. Schulz, H.~M. Krumholz, A.~L.~P. Ribeiro, and R.~Khera,
  ``Automated multilabel diagnosis on electrocardiographic images and
  signals,'' \emph{Nature communications}, vol.~13, no.~1, p. 1583, 2022.

\bibitem{c23}
X.~Guan, Y.~Lin, Q.~Wang, Z.~Liu, and C.~Liu, ``Sports fatigue detection based
  on deep learning,'' in \emph{2021 14th International Congress on Image and
  Signal Processing, BioMedical Engineering and Informatics (CISP-BMEI)}.\hskip
  1em plus 0.5em minus 0.4em\relax IEEE, 2021, pp. 1--6.

\bibitem{c24}
H.-C. Lee, C.-Y. Chen, S.-J. Lee, M.-C. Lee, C.-Y. Tsai, S.-T. Chen, and Y.-J.
  Li, ``Exploiting exercise electrocardiography to improve early diagnosis of
  atrial fibrillation with deep learning neural networks,'' \emph{Computers in
  Biology and Medicine}, vol. 146, p. 105584, 2022.

\bibitem{c25}
J.~Bi, Z.~Zhu, and Q.~Meng, ``Transformer in computer vision,'' in \emph{2021
  IEEE International Conference on Computer Science, Electronic Information
  Engineering and Intelligent Control Technology (CEI)}.\hskip 1em plus 0.5em
  minus 0.4em\relax IEEE, 2021, pp. 178--188.

\bibitem{c26}
J.~Vig and Y.~Belinkov, ``Analyzing the structure of attention in a transformer
  language model,'' \emph{arXiv preprint arXiv:1906.04284}, 2019.

\bibitem{c27}
B.~Pyakillya, N.~Kazachenko, and N.~Mikhailovsky, ``Deep learning for ecg
  classification,'' in \emph{Journal of physics: conference series}, vol. 913,
  no.~1.\hskip 1em plus 0.5em minus 0.4em\relax IOP Publishing, 2017, p.
  012004.

\bibitem{c28}
S.~W. Chen, S.~L. Wang, X.~Z. Qi, S.~M. Samuri, and C.~Yang, ``Review of ecg
  detection and classification based on deep learning: coherent taxonomy,
  motivation, open challenges and recommendations,'' \emph{Biomedical Signal
  Processing and Control}, vol.~74, p. 103493, 2022.

\bibitem{c29}
A.~Sasaki, T.~Arai, H.~Shigeta, and C.~Ibukiyama, ``Detection of silent
  myocardial ischemia patients by the spatial velocity electrocardiogram,''
  \emph{American Journal of Cardiology}, vol.~84, no.~9, pp. 1081--1083, 1999.

\bibitem{c30}
Y.~Wei, C.~Wu, G.~Li, and H.~Shi, ``Sequential transformer via an outside-in
  attention for image captioning,'' \emph{Engineering Applications of
  Artificial Intelligence}, vol. 108, p. 104574, 2022.

\bibitem{c31}
C.~Che, P.~Zhang, M.~Zhu, Y.~Qu, and B.~Jin, ``Constrained transformer network
  for ecg signal processing and arrhythmia classification,'' \emph{BMC Medical
  Informatics and Decision Making}, vol.~21, no.~1, pp. 1--13, 2021.

\bibitem{c32}
R.~Hu, J.~Chen, and L.~Zhou, ``A transformer-based deep neural network for
  arrhythmia detection using continuous ecg signals,'' \emph{Computers in
  Biology and Medicine}, vol. 144, p. 105325, 2022.

\bibitem{c33}
H.~Chefer, S.~Gur, and L.~Wolf, ``Transformer interpretability beyond attention
  visualization,'' in \emph{Proceedings of the IEEE/CVF Conference on Computer
  Vision and Pattern Recognition}, 2021, pp. 782--791.

\bibitem{c34}
G.~B. Moody and R.~G. Mark, ``The impact of the mit-bih arrhythmia database,''
  \emph{IEEE engineering in medicine and biology magazine}, vol.~20, no.~3, pp.
  45--50, 2001.

\bibitem{c35}
J.~Liu, Z.~Li, X.~Fan, X.~Hu, J.~Yan, B.~Li, Q.~Xia, J.~Zhu, and Y.~Wu,
  ``Crt-net: A generalized and scalable framework for the computer-aided
  diagnosis of electrocardiogram signals,'' \emph{Applied Soft Computing}, vol.
  128, p. 109481, 2022.

\bibitem{c36}
M.-U. Yang, D.-I. Lee, and S.~Park, ``Automated diagnosis of atrial
  fibrillation using ecg component-aware transformer,'' \emph{Computers in
  Biology and Medicine}, vol. 150, p. 106115, 2022.

\bibitem{c37}
T.-M. Chen, C.-H. Huang, E.~S. Shih, Y.-F. Hu, and M.-J. Hwang, ``Detection and
  classification of cardiac arrhythmias by a challenge-best deep learning
  neural network model,'' \emph{Iscience}, vol.~23, no.~3, p. 100886, 2020.

\bibitem{c38}
L.~Meng, W.~Tan, J.~Ma, R.~Wang, X.~Yin, and Y.~Zhang, ``Enhancing dynamic ecg
  heartbeat classification with lightweight transformer model,''
  \emph{Artificial Intelligence in medicine}, vol. 124, p. 102236, 2022.

\bibitem{c39}
G.~Yan, S.~Liang, Y.~Zhang, and F.~Liu, ``Fusing transformer model with
  temporal features for ecg heartbeat classification,'' in \emph{2019 IEEE
  International Conference on Bioinformatics and Biomedicine (BIBM)}.\hskip 1em
  plus 0.5em minus 0.4em\relax IEEE, 2019, pp. 898--905.

\bibitem{c40}
R.~Hu, J.~Chen, and L.~Zhou, ``A transformer-based deep neural network for
  arrhythmia detection using continuous ecg signals,'' \emph{Computers in
  Biology and Medicine}, vol. 144, p. 105325, 2022.

\bibitem{c41}
P.~Bing, Y.~Liu, W.~Liu, J.~Zhou, and L.~Zhu, ``Electrocardiogram
  classification using tsst-based spectrogram and convit.'' \emph{Frontiers in
  Cardiovascular Medicine}, vol.~9, pp. 983\,543--983\,543, 2022.

\bibitem{c42}
M.~D. Le, V.~S. Rathour, Q.~S. Truong, Q.~Mai, P.~Brijesh, and N.~Le,
  ``Multi-module recurrent convolutional neural network with transformer
  encoder for ecg arrhythmia classification,'' in \emph{2021 IEEE EMBS
  International Conference on Biomedical and Health Informatics (BHI)}.\hskip
  1em plus 0.5em minus 0.4em\relax IEEE, 2021, pp. 1--5.

\bibitem{c43}
\BIBentryALTinterwordspacing
 [Online]. Available:
  \url{https://www.aami.org/detail-pages/opentext-gateway-standards-committee/ec---ecg-committee}
\BIBentrySTDinterwordspacing

\bibitem{c44}
C.~Z. Hao and H.~Nugroho, ``A spatio-temporal approach with transformer network
  for heart disease classification with 12-lead electrocardiogram signals,'' in
  \emph{Control, Instrumentation and Mechatronics: Theory and Practice}.\hskip
  1em plus 0.5em minus 0.4em\relax Springer, 2022, pp. 673--684.

\end{thebibliography}

\end{document}